\def\year{2020}\relax
\documentclass[letterpaper]{article} % DO NOT CHANGE THIS
\usepackage{aaai20}  % DO NOT CHANGE THIS
\usepackage{times}  % DO NOT CHANGE THIS
\usepackage{helvet} % DO NOT CHANGE THIS
\usepackage{courier}  % DO NOT CHANGE THIS
\usepackage[hyphens]{url}  % DO NOT CHANGE THIS
\usepackage{graphicx} % DO NOT CHANGE THIS
\usepackage{graphicx}  % DO NOT CHANGE THIS

\usepackage{amsmath}
\usepackage{amssymb}
\usepackage{adjustbox}
\usepackage{multirow}
\usepackage{bm}
\usepackage[table]{xcolor}
\usepackage{CJKutf8}
\usepackage{algorithm}
\usepackage{algpseudocode}

\newcommand{\ie}{\textit{i.e.}}
\newcommand{\etc}{\textit{etc}}

\title{Enhanced Meta-Learning for Cross-lingual Named Entity Recognition with Minimal Resources}
\author{Qianhui Wu\textsuperscript{\rm 1}, Zijia Lin\textsuperscript{\rm 2}\thanks{Both authors contributed equally to this work.}, Guoxin Wang\textsuperscript{\rm 2}$^*$, Hui Chen\textsuperscript{\rm 3}, Börje F. Karlsson\textsuperscript{\rm 2}\\ \Large \textbf{Biqing Huang\textsuperscript{\rm 1}, Chin-Yew Lin\textsuperscript{\rm 2}} \\ 
	\textsuperscript{\rm 1}Beijing National Research Center for Information Science and Technology (BNRist)\\Department of Automation, Tsinghua University, Beijing
	100084, China\\
	\textsuperscript{\rm 2}Microsoft Research, Beijing, China\\
	\textsuperscript{\rm 3}Beijing National Research Center for Information Science and Technology (BNRist)\\School of Software, Tsinghua University, Beijing
	100084, China\\
	wuqianhui@tsinghua.org.cn, \{zijlin, guow, borje.karlsson, cyl\}@microsoft.com\\
	jichenhui2012@gmail.com, hbq@tsinghua.edu.cn
}

\begin{document}

\maketitle

\begin{abstract}

For languages with no annotated resources, transferring knowledge from rich-resource languages is an effective solution for named entity recognition (NER). 
While all existing methods directly transfer from source-learned model to a target language, in this paper, we propose to fine-tune the learned model with a few similar examples given a test case, which could benefit the prediction by leveraging the structural and semantic information conveyed in such similar examples.
To this end, we present a meta-learning algorithm to find a good model parameter initialization that could fast adapt to the given test case and propose to construct multiple pseudo-NER tasks for meta-training by computing sentence similarities. 
To further improve the model's generalization ability across different languages, we introduce a masking scheme and augment the loss function with an additional maximum term during meta-training. 
We conduct extensive experiments on cross-lingual named entity recognition with minimal resources over five target languages. The results show that our approach significantly outperforms existing state-of-the-art methods across the board.

\end{abstract}
\section{Introduction}
Named entity recognition (NER) is the task of locating and classifying text spans into pre-defined categories such as locations, organizations, \etc. 
It is a fundamental component in many downstream tasks. 
Most state-of-the-art NER systems employ neural architectures~\cite{huang2015bidirectional,lample2016neural,ma2016CNNBLSTMCRF,chiu2016named,peters2017semi,peters2018deep}, and thus, depend on a large amount of manually annotated data, which prevents their adaptation to low-resource languages due to the high annotation cost. 
An effective solution to this problem, which we refer to as cross-lingual named entity recognition, is transferring knowledge from a high-resource source language with abundant annotated data to a low-resource target language with limited or even no annotated data.

In this paper, we attempt to address the extreme scenario of cross-lingual transfer with minimal resources, where there is only \textit{one source language} with rich labeled data while \textit{no labeled data} is available in target languages. 
To tackle this problem, some approaches convert the cross-lingual NER task into a monolingual NER task by performing annotation projection using bilingual parallel text and word alignment information~\cite{ni2017weakly}. %,agerri2018building}. 
To eliminate the requirement of parallel texts, some methods propose to translate the labeled data of the source language at the phrase/word level, which inherently provides alignment information for label projection~\cite{mayhew2017cheap,xie2018neural}. 
% However, the resulted data are noisy because of sense ambiguity and word order differences. 
Instead of generating labeled data in target languages, other works explore language-independent features and perform cross-lingual NER in a direct-transfer manner, where the model trained on the labeled data of the source language is directly tested on target languages~\cite{tsai2016cross,ni2017weakly}. 
Among these methods, cross-lingual word representations are the most prevalent language-independent features. %wang2017amulti,huang2019cross
For example, the multilingual version of BERT~\cite{devlin2019bert} utilizes WordPiece modeling strategy to project word embeddings of different languages into a shared space and achieved state-of-the-art performance~\cite{wu2019beto}. 
In this paper, we leverage the multilingual BERT~\cite{devlin2019bert} as a base model to produce cross-lingual word representations.

While all existing direct transfer based methods straightly evaluate the source-trained model on target languages, we hold the idea that the source-trained model can be further effectively improved. Indeed, recent developments in learning cross-lingual sentence representations suggest that any sentence can be encoded into a shared space by building universal cross-lingual encoders~\cite{wu2019beto,lample2019cross}. 
By simply calculating cosine similarity between sentences in different languages with multilingual BERT~\cite{devlin2019bert}, we find that it is possible to retrieve a few source examples that are quite similar to a given target example in structure or semantics, as shown in Table \ref{tab:eg_retrieve}. 
In Example \#1, both sentences have a structure of ``Location - Date", while in Example \#2, both sentences are about people talking about sports. 
Intuitively, reviewing the structural and semantic information conveyed by similar examples might benefit prediction. 
Therefore, given a test example in a target language, we propose to first retrieve a small set of similar examples from the source language, and then, use these retrieved examples to fine-tune the model before testing.

However, if the retrieved similar set is too large, too much noise will be introduced via relatively distant examples. And thus to avoid misleading the model with distant examples, the cardinality of a similar group is typically small. 
In such a scenario, the model is expected to achieve higher performance on a test example after only one or a few fine-tuning steps using the limited-size set of retrieved examples. 
This inspires us to apply meta-learning, which aims to learn a model that facilitates fast adaptation to new tasks with a minimal amount of training examples~\cite{andrychowicz2016learning,vinyals2016matching,finn2017model}.

In this paper, we follow the recently proposed model-agnostic meta-learning approach \cite{finn2017model} and extend it to the cross-lingual NER task with minimal resources, where no labeled data is provided in target languages. 
We construct a set of pseudo-meta-NER tasks using the labeled data from the source language and propose a meta-learning algorithm to find a good model parameter initialization that could fast adapt to new tasks. 
When it comes to the adaptation phase, we regard each test example as a new task, build a pseudo training set for it, and fine-tune the meta-trained model before testing.

\begin{table}[t]
    \centering
    \setlength{\tabcolsep}{1mm}
    \begin{tabular}{c|l}
        \hline
        \multirow{2}{*}{\#1} & 
        Ginebra {\scriptsize [B-LOC]} , 23 may ( EFECOM {\scriptsize [B-ORG]} ) . \\
        &\cellcolor{gray!25}PRESS DIGEST - Israel {\scriptsize [B-LOC]} - Aug 25 .  \\
        \hline
        % \multirow{2}{*}{\#2} & 
        % Johan {\scriptsize [B-PER]} Corthouts {\scriptsize [I-PER]}\\
        % &\cellcolor{gray!25}Peter {\scriptsize [B-PER]} Smerdon {\scriptsize [I-PER]}\\
        % \hline
        \multirow{8}{*}{\#2} & 
        Flores {\scriptsize [B-PER]} afirmó: ``con él intentaremos ganar en\\
        &~~~~velocidad, que es una de las mejores virtudes que\\
        &~~~~tiene este equipo." \\
        & (Flores said: ``With him, we will try to win faster,\\
        &~~~~which is one of the best advantages of this team.")\\
        &\cellcolor{gray!25}``Things fell in for us," said Sorrento {\scriptsize [B-PER]}, who has\\
        &\cellcolor{gray!25}~~~~six career grand slams and hit the ninth of the sea-\\
        &\cellcolor{gray!25}~~~~son for the Mariners {\scriptsize [B-ORG]} .\\
        % \hline        
        % \multirow{7}{*}{\#4} & 
        % Kewell {\scriptsize [B-PER]} mantiene un conflicto con los organi-\\
        % &~~~~~~~~zadores de la gira , mientras que Viduka {\scriptsize [B-PER]}\\
        % &~~~~~~~~Moore {\scriptsize [I-PER]} y Vidmar {\scriptsize [B-PER]} están lesionados.\\
        % &(Kewell clashed with the organizers of the tour and \\
        % &~~~~~~~~Viduka Moore and Vidmar were injured.)\\
        % & \cellcolor{gray!25}Officials fed Phan {\scriptsize [B-PER]} some soup, gave him med-\\
        % & \cellcolor{gray!25}~~~~~~~~ical care, and then fed him a large breakfast.\\
        \hline
    \end{tabular}
    
    \caption{Examples of similar sentence pairs in structure (\#1) or semantics (\#2), where WHITE (\colorbox{gray!25}{GREY}) highlights the Spanish (retrieved English) examples. }
    
    \label{tab:eg_retrieve}
\end{table}

When adapting meta-learning to cross-lingual NER, we notice that most mispredictions occur on language-specific infrequent entities. 
It is known that an NER system makes predictions through word features of an entity itself, and the syntactic or semantic information of its context.
However, most entities are generally of low frequency in the training corpora of the base model, and thus, entity representations across different languages are not well-aligned in the shared embedding space.
That is, for the prediction of a low-frequency entity, over-dependence on its own features will inhibit the model transferring across languages.
Therefore, we introduce a masking scheme on named entities during meta-training to weaken the dependence on entities and promote the prediction through contextual information.
Meanwhile, considering that the commonly used average loss over all tokens treats each token equally though some tokens may be more difficult to learn and easier to be mispredict, we add a maximum term to the original loss function, which makes the model focus more on such tokens and thus reduce mispredictions, so that the meta-knowledge of these mispredictions will not be transferred to target languages.

To summarize our contributions:
\begin{itemize}
\item We propose a model-agnostic meta-learning-based approach to tackle cross-lingual NER with minimal resources. To our best knowledge, this is the first successful attempt in adapting meta-learning to NER.
\item We propose a masking scheme on named entities and augment the loss function with an additional maximum term during meta-training, to facilitate the model's ability to generalize across different languages. 
\item We evaluate our approach over 5 target languages, \ie, Spanish, Dutch, German, French, and Chinese. We show that the proposed approach significantly outperforms existing state-of-the-art methods across the board.
\end{itemize}

\section{Related Work}
\subsection{Cross-lingual NER with Minimal Resources}
There are two major branches of work in cross-lingual NER with minimal resources: methods based on annotation projection and methods based on direct transfer.

One of the typical approaches in the annotation projection category is to take bilingual parallel corpora, annotate the source side, and project the annotations to the target using learned word alignment information \cite{ni2017weakly}. %,agerri2018building}. 
However, these methods depend on parallel texts, as well as annotations in at least one side, which is unavailable in many cases.
To eliminate the requirement of parallel data, some approaches first translate source-language labeled data at the word/phrase level, and then directly copy labels across languages~\cite{xie2018neural,mayhew2017cheap}.
Yet, this might bring in too much noise due to sense ambiguity and word order differences. 
Differently, most approaches based on direct transfer leverage language-independent features to train a model on the source language and then directly apply it on target languages.
Cross-lingual word embeddings are the most widely used ones of such features~\cite{ni2017weakly,devlin2019bert}, %wang2017amulti,huang2019cross,
while other approaches also introduces word clusters~\cite{tackstrom2012} and Wikifier~\cite{tsai2016cross} as cross-lingual features. % and gazetteers \cite{zirikly2015cross} 

In this paper, we use a contextual cross-lingual word embedding~\cite{devlin2019bert} as the language-independent feature. 
Rather than directly transferring from the source-learned model to target, we propose to fine-tune the model by converting the minimal-resource cross-lingual transfer problem into a low-resource learning problem, and furthermore, present an enhanced meta-learning algorithm to tackle it. 
To our best knowledge, we are the first to extend the idea of meta-learning to cross-lingual NER with minimal resources.

\subsection{Meta-Learning}
Meta-learning has a long history \cite{naik1992meta} and emerged recently as a way to fast adapt to new tasks with very limited data.
It has been applied to various tasks such as image classification \cite{koch2015siamese,ravi2017optimization}, neural machine translation \cite{gu2018meta}, text generation \cite{huang2018natural,qian2019domain}, %,guo2019coupling}, 
and reinforcement learning \cite{finn2017model,li2018learning}

There are three categories of meta-learning algorithms:
learning a metric space which can be used to compare low-resource examples with rich-resource examples \cite{vinyals2016matching,sung2018learning}, %snell2017prototypical
learning an optimizer to update the parameters of a model \cite{andrychowicz2016learning,chen2018meta}, %munkhdalai2017meta
and learning a good parameter initialization of a model \cite{finn2017model,mi2019meta}. 

Our approach falls into the last category. We extend the idea of model-agnostic meta-learning (MAML) \cite{finn2017model} to the cross-lingual NER with minimal resources by constructing multiple pseudo-meta-NER tasks. 
Furthermore, we employ a masking scheme and enhance the loss function with an additional maximum item during meta-training to improve the model's ability to transfer across languages.

% Specifically, we take each source language training example as the query example of an individual meta task and unsupervisedly create a pseudo support set. 

% then use the fine-tuned model to make the final prediction.
% of applying the general model directly, the top-K relevant instances (in the training set) to the given test example are first selected to update the general model, which then makes the final prediction.

\section{Methodology}

Named Entity Recognition is proposed as a sequence labeling problem.
Given a sequence with $L$ tokens $\bm{x}=\{x_i\}_{i=1}^L$, an NER system is expected to produce a label sequence $\bm{y}=\{y_i\}_{i=1}^L$, where $x_i$ is the $i$-th token and $y_i$ is the corresponding label of $x_i$.
Denote the labeled training data of a source language as $D_{train}^S$ and the test data of a target language as $D_{test}^T$. 
Minimal-resource cross-lingual NER aims to train a model $\mathcal{M}$ with $D_{train}^S$ and it is expected that the model will perform well on $D_{test}^T$.

\subsection{Base Model}
In this section, we give a brief introduction to multilingual BERT \cite{devlin2019bert} (mBERT), which we leverage as the base model in our approach, since it produces an effective cross-lingual word representation. To ease the explanation, we start with BERT \cite{devlin2019bert} here.

BERT is a language model learned with the Transformer encoder \cite{Vaswani2017attention}. 
It reads the input sequence at once and learns via two strategies, \ie, masked language modeling and next sentence prediction.

mBERT follows the same model architecture and training procedure as BERT except that it is pre-trained on concatenated Wikipedia data of 104 languages. %without any cross-lingual alignment or supervision.
For tokenization, mBERT utilizes WordPiece embeddings \cite{wu2016google} with a 110k shared vocabulary to facilitate embedding space alignment across different languages. 

Following \cite{devlin2019bert} and \cite{wu2019beto}, we address cross-lingual NER by adding a linear classification layer with softmax upon the pre-trained mBERT, which can be formulated as:
\begin{equation}
    \bm{h} = \text{mBERT}(\bm{x}),
\end{equation}
\begin{equation}
    \hat{y}_l = \text{softmax}(Wh_l+b),
\end{equation}
where $\bm{x}$ is structured as $\{x_0, x_1, ..., x_L, x_{L+1}\}$. 
$x_0 = [\texttt{CLS}]$ and $x_{L+1}=[\texttt{SEP}]$ are two special tokens as in \cite{devlin2019bert}. 
$\bm{h}=\{h_l\}_{l=1}^{L}$ and $h_l$ denotes the output of the pre-trained mBERT that corresponds to the input token $x_l$. 
$\hat{y}_l$ denotes the predicted probability distribution for $x_l$. 
$W$ and $b$ are trainable parameters. 

The learning loss \textit{w.r.t.} $\bm{x}$ is modeled as the cross-entropy of the predicted label distribution and the ground-truth one for each token:
\begin{equation}
    \label{equ:original_loss}
    \mathcal{L}(\theta) = -\frac{1}{L}\sum_{l=1}^L \text{CrossEntropy}(y_l, \hat{y}_l)
\end{equation}
where $y_l$ is a one-hot vector of the ground-truth label for the $l$-th input token $x_l$. 
And the total loss for learning is the summation of losses on all training examples. 
It should be noted that, if a word is split into several subwords after tokenization, only the label of the first subword is considered. 

\subsection{Enhanced Meta-Learning for Cross-Lingual NER with Minimal Resources}
In this section, we elaborate on the proposed approach.
First, we clarify how to construct multiple pseudo-meta-NER tasks with the labeled data of the source language.
Then, we describe the meta-training algorithm of our approach.
Next, we illustrate the proposed masking mechanism and the augmented loss involved in the meta-training phase.
Finally, we show how to adapt the meta-learned model to test examples of target languages. 
The whole procedure of our algorithm is summarized in Algorithm~\ref{alg:framework}.

\subsubsection{Pseudo-Meta-NER Tasks}
In a typical meta-learning scenario, a model is trained on a set of tasks in the meta-training phase, such that the trained model can quickly adapt to new tasks using only a small number of examples.
Thus to tackle the minimal-resource cross-lingual NER via meta-learning, we first construct a set of pseudo-meta-NER tasks using the labeled data of the source language.

Assuming there are $N$ examples in $D_{train}^S = \{\bm{x}^{(i)}\}_{i=1}^N$. %, where $\bm{x}^{(i)}$ is one of them.
%Denote $\bm{x}^{(i)}$ as an example in $D_{train}^S$ and there are $N$ examples in total, \ie, $D_{train}^S = \{\bm{x}^{(i)}\}_{i=1}^N$.
We take each $\bm{x}^{(i)}$ as the test set $D_{test}^{\mathcal{T}_i}$ of an individual meta task $\mathcal{T}_i$, and create a pseudo training set $D_{train}^{\mathcal{T}_i}$ for it by retrieving the most similar examples of $\bm{x}^{(i)}$ from $D_{train}^S$. The pseudo-meta-NER tasks $\mathcal{T}_i$ can be denoted as:
\begin{equation}
    \mathcal{T}_i=(D_{train}^{\mathcal{T}_i}, D_{test}^{\mathcal{T}_i}), i \in {1, 2, ..., N}.
\end{equation}

Specifically, we first compute the sentence representation $r^{(i)}$ for each $\bm{x}^{(i)}, i \in \{1, 2, ..., N\}$: %using a pre-trained multilingual BERT~\cite{devlin2019bert}, which can be formulated as:
% \begin{equation}
%     \begin{aligned}
%     \bm{h}^{(i)} &= \text{mBERT}(\bm{x}^{(i)}),\\
%     &= \{h_{[CLS]}^{(i)}, h_1^{(i)}, h_2^{(i)}..., h_L^{(i)}, h_{[SEP]}^{(i)}\},
%     \end{aligned}
% \end{equation}
% \begin{equation}
%     r^{(i)} = h_{[CLS]}^{(i)}.
% \end{equation}
\begin{equation}
    r^{(i)} = f(\bm{x}^{(i)})
\end{equation}
where $f(\cdot)$ could be any function that is able to produce cross-lingual sentence representations. Here, we employ the multilingual BERT~\cite{devlin2019bert} and use the the final hidden vector corresponding to the first input token ([\texttt{CLS}]) as the sentence representation.

Then, we construct $D_{train}^{\mathcal{T}_i}$ by selecting top-K similar examples from $D_{train}^S \setminus \bm{x}^{(i)}$.
The metric used to measure the similarity between $\bm{x}^{(i)}$ and $\bm{x}^{(m)}$ is:
\begin{equation}
    \label{equ:similarity}
    s(\bm{x}^{(i)}, \bm{x}^{(m)}) = \frac{r^{(i)} \cdot r^{(m)}}{\Vert r^{(i)} \Vert \Vert r^{(m)} \Vert},
\end{equation}
where $m \in \{1, 2, ..., N\}$ and $m \neq i$.

\begin{algorithm}[t]
\caption{Enhanced Meta-Learning for Cross-Lingual NER with Minimal Resources}
\label{alg:framework}

\begin{algorithmic}[1]
\Procedure{Meta-Training}{$D_{train}^S$, $\alpha$, $\beta$}
    \State Construct $\mathcal{T}=\{\mathcal{T}_i\}$ with $D_{train}^S$. %\Comment{Call Recursion again}
    \State Initialize with the pre-trained base model $\mathcal{M}_{\theta}$.
    \While {not done}
        \State Sample a batch of source tasks $\mathcal{T}_i$ from $\mathcal{T}$.
        \ForAll{$\mathcal{T}_i$}
            \State Update $\theta_i' = U^n (\theta;\alpha)$. %= \arg \min_{\theta}\theta - \alpha \nabla_{\theta} \mathcal{L}_{D_{train}^{\mathcal{T}_i}}(\theta)$.
            \State Compute $g_i = \nabla_{\theta_i'} \mathcal{L}_{D_{test}^{\mathcal{T}_i}}(\theta_i')$.
        \EndFor
        \State Update $\theta \leftarrow \theta - \beta \sum_i g_i$.
    \EndWhile
    \State \textbf{return} $\mathcal{M}_{\theta^*}$ with $\theta^*$ being the final updated $\theta$
\EndProcedure
\item[]
\Procedure{Adaptation}{$\mathcal{M}_{\theta^*}$, $D_{train}^S$, $D_{test}^T$, $\gamma$}
    \ForAll{$\bm{x}^{(j)} \in D_{test}^T$}
        \State $D_{test}^{\mathcal{T}_j}=\bm{x}^{(j)}$ and construct $D_{train}^{\mathcal{T}_j}$ with $D_{train}^S$.
        \State Update $\hat{\theta}_j = \theta^* - \gamma \nabla_{\theta^*} \mathcal{L}_{D_{train}^{\mathcal{T}_j}}$.
        \State Label $D_{test}^{\mathcal{T}_j}$, \ie, $\bm{x}^{(j)}$, using $\mathcal{M}_{\hat{\theta}_j}$.
    \EndFor
\EndProcedure
\end{algorithmic}
\end{algorithm}
\subsubsection{Meta-Training}
In the meta-training phase, we train a model $\mathcal{M}$ by repeatedly simulating the \texttt{adaptation} phase, where the meta-trained model is fine-tuned with a minimal amount of training data of a new task and then tested on the test data.

Specifically, given the created pseudo-meta-NER tasks $\{\mathcal{T}_i\}_{i=1}^N$ and a model $\mathcal{M}_{\theta}$ parameterized by $\theta$, we first randomly sample a task $\mathcal{T}_i$ to derive new model parameters $\theta'$ via $n$ gradient updates on the original model parameters $\theta$, which we refer to as \texttt{inner-update}:
\begin{equation}
    % \theta_i' = \arg \min_{\theta} \mathbb{E}_{\mathcal{T}_i}\left[\mathcal{L}_{D_{train}^{\mathcal{T}_i}}\left( U^k (\theta) \right)\right]
    \theta_i' = U^n (\theta;\alpha) %= \arg \min_{\theta} \mathcal{L}_{D_{train}^{\mathcal{T}_i}}\left( \theta \right)
    \label{equ:inner_update_U}
\end{equation}
where $U^n$ is the operator that performs gradient descent %or Adam \cite{kingma2014adam} 
$n$ times with the learning rate $\alpha$ to minimize the loss $\mathcal{L}_{D_{train}^{\mathcal{T}_i}}$ computed on $D_{train}^{\mathcal{T}_i}$. For example, when applying \textit{a single gradient update},
\begin{equation}
    \label{equ:inner_update}
    \theta_i' = \theta - \alpha \nabla_{\theta} \mathcal{L}_{D_{train}^{\mathcal{T}_i}}(\theta)
\end{equation}

We then evaluate the updated parameters $\theta_i'$ on $D_{test}^{\mathcal{T}_i}$ and further update the meta model $\mathcal{M}_{\theta}$ by minimizing the loss $\mathcal{L}_{D_{test}^{\mathcal{T}_i}} (\theta_i')$ with respect to $\theta$, which is referred to as \texttt{meta-update}. When aggregating multiple pseudo-meta-NER tasks, the meta-objective is: 
\begin{equation}
\begin{aligned}
    \min_{\theta}  \sum_i \mathcal{L}_{D_{test}^{\mathcal{T}_i}} (\theta_i')
\end{aligned}
\end{equation}
Take \textit{a single gradient update} with the learning rate $\beta$, the meta-update can be formulated as:
\begin{equation}
    \label{equ:meta_update}
    \begin{aligned}
    \theta &\leftarrow \theta - \beta \sum_i \nabla_{\theta} \mathcal{L}_{D_{test}^{\mathcal{T}_i}} (\theta_i') \\
    &= \theta - \beta \sum_i g_i
    \end{aligned}
\end{equation}
where $g_i$ is the meta-gradient on task $\mathcal{T}_i$, which can be expanded to:
\begin{equation}
% \begin{aligned}
    \label{equ:meta_gradient}
    g_i = \nabla_{\theta}\mathcal{L}_{D_{test}^{\mathcal{T}_i}}(\theta_i')= \nabla_{\theta_i'}\mathcal{L}_{D_{test}^{\mathcal{T}_i}}(\theta_i') \nabla_{\theta}(\theta_i')
% \end{aligned}
\end{equation}

In Equation \ref{equ:meta_gradient}, $\nabla_{\theta}(\theta_i')$ is the Jacobian matrix of the update operation $U^n$ that will introduce higher order gradient.
To reduce computational cost, we use a first-order approximation by replacing the Jacobian $\nabla_{\theta}(\theta_i')$ with the identity matrix as in~\cite{finn2017model}. % \citeauthor{finn2017model}~\shortcite{finn2017model}.
Therefore, $g_i$ can be computed as:
\begin{equation}
    g_i = \nabla_{\theta_i'}\mathcal{L}_{D_{test}^{\mathcal{T}_i}}(\theta_i')
\end{equation}

% The entire training procedure can be summarized in Algorithm \ref{alg:meta_train}. 
Compared with the common training scheme, the meta-learned model is more sensitive to the changes among different tasks, which can promote the learning of the common internal representations rather than the distinctive features of the source language training data $D_{train}^S$.
When coming to the adaptation phase, the model could be more sensitive to the features of new tasks, and hence only one or a few fine-tune epochs on a minimal amount of data can make rapid progress without overfitting~\cite{finn2017model}.

\subsubsection{Masking on Named Entities} 
For cross-lingual NER with minimal resources, the alignments of the entity representations in the shared space are particularly important as this task focuses on understanding entities across languages. %in a target language with the information learned from the source language.
However, compared with commonly used words, most entities are of low frequency in the pre-training corpora of the base model. 
As a result, the learned entity representations across languages are not well-aligned in the shared space. 

In order to reduce the dependence on target entity representations and encourage the model to predict through context information, we employ the [\texttt{MASK}] token as introduced in~\cite{devlin2019bert} to mask entities at the token level in each training example, \ie, each token inside an entity is randomly masked with a given probability. Then, the masked examples are fed as input data for the model.
Note that we re-perform the masking scheme at the beginning of each training epoch.

\subsubsection{Max Loss}
In Equation~\ref{equ:original_loss}, the loss for each token is uniformly weighted so that all tokens contribute equally when training the model. 
Nonetheless, this will result in insufficient learning for those tokens with relatively higher losses. 
In order to force the model to put more effort in learning from such tokens, we modify the loss function as:
\begin{equation}
\begin{aligned}
    \label{equ:with_max_loss}
    \mathcal{L}(\theta) = & -\frac{1}{L}\sum_{i=1}^L 
    \text{CrossEntropy}(y_i, \hat{y}_i) \\
    & - \lambda \max_{i\in\{1, 2, ..., L\}}
    \text{CrossEntropy}(y_i, \hat{y}_i) 
    %y_i\log (\hat{y}_i) - \lambda \max_{i\in\{1, 2, ..., L\}}y_i\log (\hat{y_i}).
\end{aligned}
\end{equation}
where $\lambda \geq 0$ is a weighting factor. In this way, the potential mispredictions of the high-loss tokens would probably be corrected during meta-training. 
The benefit of such correction is that the meta-knowledge about the mispredictions, which is also going to be transferred to target tasks, would be reduced, so that the model could achieve better performance after transferring.

In summary, the $\mathcal{L}_{D_{train}^{\mathcal{T}_i}}$ and $\mathcal{L}_{D_{test}^{\mathcal{T}_i}}$ in Meta-Training of Algorithm~\ref{alg:framework} are with the masking scheme and the max loss.
\subsubsection{Adaptation}
When it comes to the adaptation phase, \ie, applying $\mathcal{M}_{\theta^*}$ on target languages, we take each test example $\bm{x}^{(j)} \in D_{test}^T$ as the test set $D_{test}^{\mathcal{T}_j}$ of a target task $\mathcal{T}_j$. 
We then construct a pseudo training set $D_{train}^{\mathcal{T}_j}$ for each $\mathcal{T}_j$ by retrieving top-K similar examples of $\bm{x}^{(j)}$ from the source language training data $D_{train}^S$ using the metric in Equation \ref{equ:similarity}. 
Subsequently, we fine-tune the meta-learned model $\mathcal{M}_{\theta^*}$ with the pseudo training set $D_{train}^{\mathcal{T}_j}$ as in Equation~\ref{equ:original_loss}  via \textit{one gradient update}, and then use the fine-tuned model to predict labels for the test set $D_{test}^{\mathcal{T}_j}$, \ie, $\bm{x}^{(j)}$. 

It should be noted that in the adaptation phase, we do not perform the masking scheme to avoid information loss of target entities. 
Besides, since the size of the pseudo training set is very small, we employ the loss function as in Equation~\ref{equ:original_loss} rather than Equation~\ref{equ:with_max_loss} to prevent over-adjusting on uncertain or mispredicted tokens. 
In fact, when using Equation~\ref{equ:with_max_loss} for adaptation, the model could achieve slightly better performance in some cases but also get worse performance in others due to the mentioned over-adjustment.
\section{Experiments}
In this section, we evaluate our enhanced meta-learning approach for cross-lingual NER with minimal resources and compare our approach to current state-of-the-art methods.

\subsection{Datasets}
We conduct experiments on four benchmark datasets: CoNLL-2002 Spanish and Dutch NER \cite{tjong2002introduction}, CoNLL-2003 English and German NER \cite{tjong2003introduction},  Europeana Newspapers French NER \cite{neudecker2016corpus}, and MSRA Chinese NER \cite{cao2018adversarial}. 
Table~\ref{tab:dataset} shows the statistics of all datasets. 

\begin{itemize}
    \item \textbf{CoNLL-2002/2003} is annotated with four entity types: \texttt{PER}, \texttt{LOC}, \texttt{ORG}, and \texttt{MISC}. All datasets are split into a training set, a development set (testa) and a test set (testb).
    
    \item \textbf{Europeana Newspapers} is annotated with three types: \texttt{PER}, \texttt{LOC}, and \texttt{ORG}. We randomly sample $10\%$ of sentences from the whole data to build a test set.
    
    \item \textbf{MSRA} is also annotated with three types: \texttt{PER}, \texttt{LOC}, and \texttt{ORG}. Since gold word segmentation is not provided in the test set, we use word segmentation from \cite{zhang2018chinese}. %\citeauthor{zhang2018chinese}~\shortcite{zhang2018chinese}.
\end{itemize}

\begin{table}[t]
    \centering
    \setlength{\tabcolsep}{1.5mm}
    \scalebox{0.85}{
    \begin{tabular}{|c|c|c|c|c|}
      \hline
      Language & Type &Train &Dev &Test  \\ \hline
      English-en &Sentence &14,987 &3,466 &3,684 \\
      (CoNLL-2003)&Entity &23,499 &5,942 &5,648 \\
      \hline
      German-de &Sentence &12,705 &3,068 &3,160 \\
      (CoNLL-2003)&Entity &11,851 &4,833 &3,673 \\
      \hline
      Spanish-es &Sentence &8,323 &1,915 &1,517 \\
      (CoNLL-2002)&Entity &18,798 &4,351 &3,558 \\
      \hline
      Dutch-nl &Sentence &15,806 &2,895 &5,195 \\
      (CoNLL-2002)&Entity &13,344 &2,616 &3,941 \\
      \hline
      French-fr &Sentence &9,527 & - &2,375 \\
      (Europeana News)&Entity &7,899 & - &1,984 \\
      \hline
      Chinese-zh &Sentence &46,306 & - &4,361 \\
      (MSRA)&Entity &72,864 & - &5,478 \\
      \hline
    \end{tabular}}
    \caption{Dataset statistics.}
    \label{tab:dataset}
\end{table}

For all experiments, we use English as the source language and the others as target languages, \ie, the model $\mathcal{M}$ is trained on the training set of English data and evaluated on the test sets of each other language. 
When transferring to French and Chinese, we relabel the \texttt{MISC} entities in English training data into non-entities for meta-training as there is no \texttt{MISC} in the French and Chinese test sets.
Following~\cite{wu2019beto}, we use the BIO labeling scheme.

\begin{table*}[t]
  \centering
    \begin{adjustbox}{max width=0.75\textwidth}
    \begin{tabular}{c|c|c|c|c|c|c}
      \hline
         &	es	&	nl	&	de	&	fr	&	zh	&	Average	\\ \hline
        \citeauthor{tackstrom2012}~\shortcite{tackstrom2012}&	59.30	&	58.40	&	40.40	&	-	&	-	&	-	\\ \hline
        \citeauthor{tsai2016cross}~\shortcite{tsai2016cross}&	60.55	&	61.56	&	48.12	&	-	&	-	&	-	\\ \hline
        \citeauthor{ni2017weakly}~\shortcite{ni2017weakly}&	65.10	&	65.40	&	58.50	&	-	&	-	&	-	\\ \hline
        \citeauthor{mayhew2017cheap}~\shortcite{mayhew2017cheap}&	65.95	&	66.50	&	59.11	&	-	&	-	&	-	\\ \hline
        \citeauthor{xie2018neural}~\shortcite{xie2018neural}&	72.37	&	71.25	&	57.76	&	-	&	-	&	-	\\ \hline
        \citeauthor{wu2019beto}~\shortcite{wu2019beto}&	74.96	&	77.57	&	69.56	&	-	&	-	&	-	\\ \hline
        Base Model &	74.59	&	79.57	&	70.79	&	50.89	&	76.42	&	70.45\\ \hline %mBERT \cite{devlin2019bert} (

        \textbf{Ours} &	\textbf{76.75}	&	\textbf{80.44}	&	\textbf{73.16}	&	\textbf{55.30}	&	\textbf{77.89}	&	\textbf{72.71} \\ \hline
    \end{tabular}
    \end{adjustbox}
    \caption{Results of cross-lingual NER with minimal resources\footnotemark.} %\textcolor{red}{``*"}} indicate statistically significant difference ($p<0.01$) from the baseline method mBERT~\cite{devlin2019bert}}.}
    \label{tab:zero_shot}
\end{table*}

% \begin{table*}[t]
%   \centering
%     \begin{adjustbox}{max width=\textwidth}
%     \begin{tabular}{l|c|c|c|c|c|c}
%       \hline
%          &	es	&	nl	&	de	&	fr	&	zh	&	Average	\\ \hline
%         mBERT	&	74.63	&		79.46	&	70.09	&	51.08	&	76.19	&	70.29 \\ \hline
%         ~~~~ + [fine-tune] & 74.56	&	79.49	&	70.26	&	49.95	&	75.31 & 69.91\\ \hline

%     \end{tabular}
%     \end{adjustbox}
%     \caption{Zero-shot cross-lingual NER F1 scores of mBERT with and without fine-tune.}
%     \label{tab:ablation_meta}
% \end{table*}
% 为 zero-shot 和 low-resource 分开挑 seed 可以取得更好看的结果
\begin{table*}[t]
  \centering
    \setlength{\tabcolsep}{1mm}
    \begin{adjustbox}{max width=0.95\textwidth}
    \begin{tabular}{l|c|c|c|c|c|c}
      \hline
         &	es	&	nl	&	de	&	fr	&	zh	& Average\\ \hline
        \textbf{Ours} &	\textbf{76.75}	&	\textbf{80.44}	&	\textbf{73.16}	&	\textbf{55.30}	&	\textbf{77.89} &  \textbf{72.71} \\ \hline
        Ours w/o max loss &76.05	(-0.70)	&	79.50	(-0.94)	&	71.84	(-1.32)	&	52.64	(-2.66)	&	76.77	(-1.12) & 71.36 (-1.35)\\ \hline
        Ours w/o masking	&	75.57	(-1.18)	&	80.38	(-0.06)	&	72.76	(-0.40)	&	54.29	(-1.01)	&	77.79	(-0.10) & 72.16 (-0.55)\\ \hline
        Ours w/o max loss/masking &	75.33	(-1.42)	&	80.13	(-0.31)	&	71.49	(-1.67)	&	52.79	(-2.51)	&	76.50	(-1.39) & 71.25 (-1.46)\\ \hline
        % Ours w/o meta-train/max loss/masking (\ie, Base Model)&74.59	(-2.16)	&79.57	(-0.87)&70.79	(-2.37)&50.89	(-4.41)	&76.42	(-1.47) &70.45 (-2.26)\\
        Ours w/o meta-train/max loss/masking & \multirow{2}{*}{74.59	(-2.16)}	&	\multirow{2}{*}{79.57	(-0.87)}	&	\multirow{2}{*}{70.79	(-2.37)}	&	\multirow{2}{*}{50.89	(-4.41)}	&	\multirow{2}{*}{76.42	(-1.47)} & \multirow{2}{*}{70.45 (-2.26)} \\
        \multicolumn{1}{c|}{(\ie, Base Model)} & & & & & \\
        \hline
    \end{tabular}
    \end{adjustbox}
    \caption{Ablation study on cross-lingual NER with minimal resources.} % where ``$-$" denotes removing the corresponding strategies.}
    \label{tab:zero_shot_ablation}
\end{table*}

\subsection{Implementation Details}
We implement our approach with PyTorch 1.0.1.
We use the cased multilingual $\text{BERT}_{\text{BASE}}$ with 12 Transformer blocks, 768 hidden units, 12 self-attention heads, GELU activations~\cite{dan2016bridging}, a dropout rate of 0.1 and learned positional embeddings. We employ WordPiece embeddings~\cite{wu2016google} to split a word into subwords, which are then directly fed into the model without any other pre-processing. 
We empirically select the hyper-parameters and utilize them in all experiments.
Specifically, for sequence length, we employ a sliding window with a maximum length of $128$.
When the sequence length is larger than $128$, the last $64$ subwords of the first window are kept as the context for the subsequent window. 
Following~\cite{huang2018natural}, we select $K=2$ similar examples for both pseudo NER task construction and the adaptation phase.
The mask ratio is set to $0.2$, $\lambda$ in Equation~\ref{equ:with_max_loss} is set to $2.0$, update steps $n$ in Equation~\ref{equ:inner_update_U} is set to $2$, the number of sampled pseudo-NER tasks used for one meta-update is set to $32$, and the maximum meta-update steps is set to $3\times 10^3$. 
Following~\cite{wu2019beto}, we freeze the parameters of the embedding layer and the bottom three layers of the base model. 
According to the suggestions of model hyper parameters in~\cite{devlin2019bert}, for the optimizers of both inner-update and meta-update, we use Adam~\cite{kingma2014adam} with learning rate of $\alpha, \beta=3e-5$, %$\beta_1$ of $0.9$, $\beta_2$ of $0.999$, and L2 weight decay of $0.01$, 
while for gradient updates during adaptation, we set the learning rate $\gamma$ to 1e-5. 
% Additionally, we exploit learning rate warm-up over the first 300 meta-update steps and linear decay of the learning rate. 
Following \cite{tjong2002introduction}, we use the phrase level F1-score as the evaluation metric. 
To reduce the model bias, we carry out $5$ runs and report the average performance. 

% \subsection{\textcolor{red}{Baselines}}
% Here, we compared our approach with various state-of-the-art methods, including direct transfer and translation based label projection.
% As shown in Table \ref{tab:zero_shot}, \cite{tackstrom2012}, \cite{tsai2016cross}, and \cite{ni2017weakly} proposed to use language-independent features such as cross-lingual word clusters and wikifier features to perform zero-shot cross-lingual in a manner of direct transfer.
% \cite{mayhew2017cheap} and \cite{xie2018neural} first translated the source language training data into target languages and trained a NER model for each target language.
% \cite{wu2019beto} is the most related work that makes use of mBERT to produce cross-lingual word embeddings.
% Since we take the pre-trained mBERT~\cite{devlin2019bert} as the base model, we also perform zero-shot cross-lingual NER using it with the same experimental settings as in our approach and regard the results as the baselines.
% We also provide the results of the base model with completely the same experimental settings used in our approach. 

\subsection{Performance Comparison}
Table~\ref{tab:zero_shot} presents our results on transferring from English to five other languages, alongside results from previous works. 
The results show that our approach significantly outperforms the previous state-of-the-art methods across the board, with relative improvements on F1-score compared to the base model ranging from $1.09$\% for Dutch to $8.67$\% for French (with average improvement of $3.21$\%), which demonstrates the effectiveness of the proposed enhanced meta-learning algorithm. \footnotetext{The \texttt{zh} results reported in \cite{wu2019beto} used a dataset not specified in the paper, so we don't list them here.}

Particularly, compared with the base model, our approach achieves particularly significant improvement on German and French, which can be attributed to our model's stronger ability to predict through context information. 
In English, proper nouns of \texttt{LOCATION}, \texttt{PERSON}, \etc. often begin with a capital letter while most general nouns do not. 
As a result, without effective extraction of context information, the base model tends to mislabel capitalized terms for general nouns as entities, and such phenomenon is especially serious when adapting the model to French and German, where capitalization rules differ from English for general nouns, titles, \etc. or due to noise in datasets. 
In contrast, our approach is more robust in such cases due to the introduction of the masking scheme and the max loss, which facilitates the model to label general nouns as non-entities based more on context.

\subsection{Ablation Study}
We propose several strategies to enhance the base model, including the meta-training and adaptation procedure, the masking scheme, and the augmented loss.
In this section, we conduct ablation study experiments to investigate the influence of these factors. Table \ref{tab:zero_shot_ablation} shows the results. 

\begin{itemize}
    \item \textit{Ours w/o max loss}, which removes the additional maximum term in the loss function. The performance in terms of F1-score decreases by $1.35$ on average. We conjecture that, without the maximum term, the meta-knowledge from mispredictions is transferred to target tasks along with the meta-model, which hurts performance.
    \item \textit{Ours w/o masking}, which wipes out the masking scheme during the meta-training phase. This causes a performance drop across all languages, with a maximum drop of $1.18$ F1-score in Spanish. That further demonstrates the necessity of predicting through contextual information.
    \item \textit{Ours w/o max loss/masking}, which cuts out both the masking scheme and the max loss at once. In this case, our approach degrades into the base model trained with merely model-agnostic meta-learning. This results in a performance drop of $1.46$ F1-score on average, indicating that both the masking scheme and the max loss do bring enhancement to meta-learning for cross-lingual NER with minimal resources.
    \item \textit{Ours w/o meta-train/max loss/masking}, which further eliminates the meta-training and the adaptation phase from \textit{Ours w/o max loss/masking}. In that case, our approach degenerates into the base model.%, \ie, the multilingual BERT~\cite{devlin2019bert}. 
    From Table \ref{tab:zero_shot_ablation}, we can see that this will lead to a significant and consistent performance drop on all five target languages, which demonstrates the effectiveness of meta-leaning employed in our approach.
\end{itemize}

\begin{table*}[t]
  \centering
    \begin{adjustbox}{max width=0.88\textwidth}
    \begin{tabular}{|c|l|}
      \hline
        %  \textbf{\#1}{ [Spanish]: }\colorbox[rgb]{1, 1, 1}{... Fidalgo was elected to replace Antonio Gutierrez at the head of the General Secretariat.}\\
        %  \hdashline
        \textbf{\#1} & \textbf{Base Model: }... \colorbox[rgb]{0.80, 0.98, 0.85}{[{\scriptsize PER} Fidalgo]} fue elegido para sustituir a \colorbox[rgb]{0.80, 0.98, 0.85}{[{\scriptsize PER} Antonio Gutiérrez]} al frente de la \colorbox[rgb]{0.97, 0.82, 0.80}{Secretaría General}.\\
        % \cline{2-2}
        \multirow{2}{*}{Spanish} & \textbf{Ours: }...  \colorbox[rgb]{0.80, 0.98, 0.85}{[{\scriptsize PER} Fidalgo]} fue elegido para sustituir a \colorbox[rgb]{0.80, 0.98, 0.85}{[{\scriptsize PER} Antonio Gutiérrez]} al frente de la \colorbox[rgb]{0.80, 0.98, 0.85}{[{\scriptsize ORG} Secretaría General]}.\\
        % \cline{2-2}
        & \textbf{Translation in English: }\colorbox[rgb]{1, 1, 1}{... Fidalgo was elected to replace Antonio Gutierrez at the head of the General Secretariat.}\\
        \hline
        \hline
         
        \textbf{\#2} & \textbf{Base Model: }... hebben politie het speelplein \colorbox[rgb]{0.80, 0.98, 0.85}{[{\scriptsize LOC} Redoute]} aan de \colorbox[rgb]{0.97, 0.82, 0.80}{[{\scriptsize PER} Edmond Thieffrylaan]} schoongeveegd van junkies.\\
        % \cline{2-2}
        \multirow{2}{*}{Dutch} & \textbf{Ours: }... hebben politie het speelplein \colorbox[rgb]{0.80, 0.98, 0.85}{[{\scriptsize LOC} Redoute]} aan de \colorbox[rgb]{0.80, 0.98, 0.85}{[{\scriptsize LOC} Edmond Thieffrylaan]} schoongeveegd van junkies.\\
        % \cline{2-2}
        & \textbf{Translation in English: }\colorbox[rgb]{1, 1, 1}{... the police have cleaned up the Redoute playground on Edmond Thieffrylaan from junkies.}\\
        \hline
        \hline
         
        \textbf{\#3} & \textbf{Base Model: }
        Kurzfristig wurde der Vorgänger \colorbox[rgb]{0.97, 0.82, 0.80}{[{\scriptsize ORG} Krauses]} , \colorbox[rgb]{0.80, 0.98, 0.85}{[{\scriptsize PER} Peter Wünsch]}, zurückgeholt.\\
        % \cline{2-2}
        \multirow{2}{*}{German} & \textbf{Ours: }
        Kurzfristig wurde der Vorgänger \colorbox[rgb]{0.80, 0.98, 0.85}{[{\scriptsize PER} Krauses]} , \colorbox[rgb]{0.80, 0.98, 0.85}{[{\scriptsize PER} Peter Wünsch]}, zurückgeholt.\\
        % \cline{2-2}
        & \textbf{Translation in English: }\colorbox[rgb]{1, 1, 1}{Not long ago, Krauses' predecessor, Peter Wünsch, was taken back.}\\
        \hline
        \hline
         
        % \textbf{\#4} & \textbf{Base Model: }A 2 heures , route de \colorbox[rgb]{0.97, 0.82, 0.80}{[{\scriptsize PER} Landivisiau]}, course au galop le prix, 30 fr.\\
        % % \cline{2-2}
        % \multirow{2}{*}{French} & \textbf{Ours: }A 2 heures , route de \colorbox[rgb]{0.80, 0.98, 0.85}{[{\scriptsize LOC} Landivisiau]}, course au galop le prix, 30 fr.\\
        % % \cline{2-2}
        % & \textbf{Translation in English: }\colorbox[rgb]{1, 1, 1}{At 2 o'clock, road of Landivisiau, galloping race the price, 30 fr.}\\
        %  \hline
        %  \hline
        
        \textbf{\#4} & \textbf{Base Model: }
        \begin{CJK}{UTF8}{gbsn}
        在 \colorbox[rgb]{0.80, 0.98, 0.85}{[{\scriptsize LOC} 希腊]} ， 人们 常称 他 为 `` 小 \colorbox[rgb]{0.97, 0.82, 0.80}{奥纳西斯} " 。
        \end{CJK}
        \\
        % \cline{2-2}
        \multirow{2}{*}{Chinese} & \textbf{Ours: }
        \begin{CJK}{UTF8}{gbsn}
        在 \colorbox[rgb]{0.80, 0.98, 0.85}{[{\scriptsize LOC} 希腊]} ， 人们 常称 他 为 `` 小 \colorbox[rgb]{0.80, 0.98, 0.85}{[{\scriptsize PER} 奥纳西斯]} " 。
        \end{CJK}
        \\
        % \cline{2-2}
        & \textbf{Translation in English: }\colorbox[rgb]{1, 1, 1}{In Greece, he is often called "Little Onassis".}\\
         
        %  \textbf{\#5 [Chinese(\textcolor{red}{Replace to Another case})]: }\colorbox[rgb]{1, 1, 1}{He said that Huangpu Military Academy was founded by Dr. Sun Yat-sen, the pioneer of the Democratic}\\
        %  \colorbox[rgb]{1, 1, 1}{revolution, and the Communist Party of China represented by Zhou Enlai.}\\
        %  \hdashline
        %  \textbf{Base Model: }
        %  \begin{CJK}{UTF8}{gbsn}
        %  他 说 ， \colorbox[rgb]{0.80, 0.98, 0.85}{[{\scriptsize ORG} 黄埔军校]} 是 民主革命 的 先行者 \colorbox[rgb]{0.80, 0.98, 0.85}{[{\scriptsize PER} 孙中山]} 先生 和 以 \colorbox[rgb]{0.80, 0.98, 0.85}{[{\scriptsize PER} 周恩来]} 为 代表 的
        %  \end{CJK}\\
        %  \begin{CJK}{UTF8}{gbsn}
        %  \colorbox[rgb]{0.97, 0.82, 0.80}{中国共产党} 人 一道 创建 的 。
        %  \end{CJK}\\
        %  \hdashline
        %  \textbf{Ours: }
        %  \begin{CJK}{UTF8}{gbsn}
        %  他 说 ， \colorbox[rgb]{0.80, 0.98, 0.85}{[{\scriptsize ORG} 黄埔军校]} 是 民主革命 的 先行者 \colorbox[rgb]{0.80, 0.98, 0.85}{[{\scriptsize PER} 孙中山]} 先生 和 以 \colorbox[rgb]{0.80, 0.98, 0.85}{[{\scriptsize PER} 周恩来]} 为 代表 的
        %  \end{CJK}\\
        %  \begin{CJK}{UTF8}{gbsn}
        %  \colorbox[rgb]{0.80, 0.98, 0.85}{[{\scriptsize ORG} 中国共产党]} 人 一道 创建 的 。
        %  \end{CJK}\\
    \hline
    
    \end{tabular}
    \end{adjustbox}
    \caption{Case study of cross-lingual NER with minimal resources. The \colorbox[rgb]{0.80, 0.98, 0.85}{GREEN} (\colorbox[rgb]{0.97, 0.82, 0.80}{RED}) highlight indicates a correct (incorrect) label.}
    \label{tab:case_study}
\end{table*}
\subsection{Case Study}
We give a case study to analyze the quality of the results produced by our approach and the base model. Table~\ref{tab:case_study} demonstrates that our approach has a stronger ability to transfer semantic information.

In example \#1, the base model fails to identity ``Secretar\'ia General" as \texttt{ORG}, probably because its most similar phrase ``secretary general" in the English dataset is usually labeled as non-entities. 
% all of which is labeled with \texttt{O}s in the English dataset such as ``quit[\texttt{O}] as[\texttt{O}] Sakigake[\texttt{B-ORG}] secretary[\texttt{O}] general[\texttt{O}]". 
However, our approach can recognize it according to the learned semantic information ``a \texttt{PER} was selected to replace another \texttt{PER} at the head of an \texttt{ORG}". 
Similarly, in example \#2, the base model incorrectly labels ``Edmond Thieffrylaan" as \texttt{PER}. We suspect that this is because Edmond appears as a part of a person name ``Jim Edmond" in the English training data. Surprisingly, the proposed approach labels it as \texttt{LOC} correctly according to the context ``clean up the playground on \texttt{LOC}". 
Moreover, the baseline model mispredicts the labels of ``Krauses" in Example \#3 and \begin{CJK}{UTF8}{gbsn}``奥纳西斯"\end{CJK}
in Example \#4, two unseen entities in the English training data, while our approach gives the right prediction on the basis of context information. 
Considering the limited space, we provide more cases in Table S1 of the supplementary material. %~\ref{tab:sup_case_study}

\begin{table}[t]
  \centering
	\setlength{\tabcolsep}{1mm}
    \scalebox{0.9}{
    \begin{tabular}{c|c|c|c|c|c|c|c}
        \hline
        data & systems &	es	&	nl	&	de	&	fr	&	zh	&	Average	\\ 
         \hline
        \multirow{5}{*}{1\%} & CL &	60.43	&	54.77	&	-	&	-	&	-	&	-	\\
        &ML&	62.72	&	63.57	&	-	&	-	&	-	&	- \\ 
        &MLMT &	68.33	&	66.73	&	-	&	-	&	-	&	- \\ 
        &Base Model &	76.83	&	80.90	&	73.22	&	60.51	&	79.72	&	74.24 \\
        &\textbf{Ours} &	 \textbf{78.59}	&	\textbf{82.72}	&	\textbf{75.12}	&	\textbf{61.71}	&	\textbf{80.34}	&	\textbf{75.70} \\ 
        \hline
        \multirow{5}{*}{2\%} & CL &	66.45	&	61.91	&	-	&	-	&	-	&	-\\
        &ML&	71.22	&	70.62	&	-	&	-	&	-	&	- \\ 
        &MLMT &	72.59	&	70.92	&	-	&	-	&	-	&	- \\ 
        &Base Model &	77.28	&	81.54	&	74.31	&	64.43	&	81.86	&	75.88 \\
        &\textbf{Ours} &	 \textbf{79.54}	&	\textbf{83.07}	&	\textbf{75.64}	&	\textbf{65.79}	&	\textbf{82.58}	&	\textbf{77.32}\\ 
        \hline
        \multirow{5}{*}{5\%} & CL &	69.63	&	70.29	&	-	&	-	&	-	&	-\\
        &ML&	76.15	&	76.52	&	-	&	-	&	-	&	-\\ 
        &MLMT &	77.13	&	77.03	&	-	&	-	&	-	&	-\\ 
        &Base Model &78.74	&	82.00	&	75.48	&	67.73	&	85.11	&	77.81\\
        &\textbf{Ours} &	\textbf{80.32}	&	\textbf{83.72}	&	\textbf{77.70}	&	\textbf{69.19}	&	\textbf{85.67}	&	\textbf{79.32}\\ 
        \hline
    \end{tabular}}
    
    % \end{adjustbox}
    
    \caption{Low resource cross-lingual NER results, where x\% denotes the percentage of labeled training data in target languages used in the adaptation phase.\\ \textbf{CL}: Cross-lingual transfer using a shared character embedding layer~\cite{yang2017transer}. \\
    \textbf{ML}: The multi-lingual framework as in \citeauthor{lin2018amulti}~\shortcite{lin2018amulti}. \\
    \textbf{MLMT}: The multi-lingual multi-task framework as in \citeauthor{li2018learning}~\shortcite{li2018learning}.}
    %\textbf{Base Model}: the multi-lingual BERT as in \citeauthor{devlin2019bert}~\shortcite{devlin2019bert}.} %\textcolor{red}{``*" indicate statistically significant difference ($p<0.01$) from the baseline method Base Model~\cite{devlin2019bert}}.}
    \label{tab:few_shot}
\end{table}

\subsection{Discussion: Extend to Low-Resource Cross-Lingual NER}
Here, we extend the proposed approach to the task of low-resource cross-lingual NER.
To simulate a low-resource setting, we use randomly sampled subsets of the training data of a target language.
Compared with minimal-resource cross-lingual transfer, we take the same meta-training procedure.
For the adaptation phase, we directly use the entire subsets to fine-tune the meta-learned model for efficiency, and then test on the test data of the target language.

We compare our meta-learning based approach with other multi-lingual and multi-task based approaches.
For the results not reported in \cite{yang2017transer} and \cite{lin2018amulti}, we re-implement their methods based on the open-source github repositories\footnote{https://github.com/kimiyoung/transfer}$^\text{,}$\footnote{https://github.com/limteng-rpi/mlmt}$^\text{,}$\footnote{We re-implement only Spanish and Dutch as the original repositories only provide aligned word embeddings for these two languages.}.
As presented in Table \ref{tab:few_shot}, our approach significantly outperforms other approaches across all target languages with different percentage of labeled data. 
Compared with the base model, there is an average improvement of 1.47 F1-score. 
We also study the factor analysis of the enhanced meta-learning algorithm under low-resource setting. 
One can refer to Table S2 of the supplementary material for details due to the limited space. %\ref{tab:sup_ablation}
Similarly, removing any factor in our proposed approach will lead to a performance drop, which further demonstrates that our approach is reasonable.
\section{Conclusion}
In this paper, we propose an enhanced meta-learning algorithm for cross-lingual NER with minimal resources, considering that the model could achieve better results after a few fine-tuning steps over a very limited set of structurally/semantically similar examples from the source language.
To this end, we propose to construct multiple pseudo-NER tasks for meta-training by computing sentence similarities.
Moreover, in order to improve the model’s capability to transfer across different languages, we present a masking scheme and augment the loss function with an additional maximum term during meta-training.
Experiments on five target languages show that the proposed approach leads to new state-of-the-art results with a relative F1-score improvement of up to 8.76\%. 
We also extend the approach to low-resource cross-lingual NER, and it also achieves state-of-the-art results.

\bibliographystyle{aaai}
\bibliography{refs}

% \beginsupplement

% \input{sections/supplementary.tex}

\end{document}